\documentclass[aps,prapplied,twocolumn,amsmath,amssymb]{revtex4-2}

\usepackage{float}
\usepackage{graphicx}
\usepackage{dcolumn}
\usepackage{bm}

\begin{document}


\title{High-fidelity Prediction of Megapixel Longitudinal Phase-space Images of Electron Beams using Encoder-Decoder Neural Networks}


\author{J. Zhu, Y. Chen, F. Brinker, W. Decking, S. Tomin, H. Schlarb}
\affiliation{Deutsches Elektronen-Synchrotron DESY, Notkestrasse 85, 22607 Hamburg, Germany}


\date{\today}

\begin{abstract}
Modeling of large-scale research facilities is extremely challenging due to complex physical processes and engineering problems. Here, we adopt a data-driven approach to model the longitudinal phase-space diagnostic beamline at the photoinector of the European XFEL with an encoder-decoder neural network model. A deep convolutional neural network (decoder) is used to build images measured on the screen from a small feature map generated by another neural network (encoder). We demonstrate that the model trained only with experimental data can make high-fidelity predictions of megapixel images for the longitudinal phase-space measurement without any prior knowledge of photoinjectors and electron beams. The prediction significantly outperforms existing methods. We also show the scalability and interpretability of the model by sharing the same decoder with more than one encoder used for different setups of the photoinjector, and propose a pragmatic way to model a facility with various diagnostics and working points. This opens the door to a new way of accurately modeling a photoinjector using neural networks and experimental data. The approach can possibly be extended to the whole accelerator and even other types of scientific facilities. 
\end{abstract}


\maketitle


\section{INTRODUCTION}

Operations of large-scale scientific user facilities like the  European XFEL \cite{Decking2020} are very challenging as it is required to meet the specifications of various user experiments \cite{Pascarelli2020} and be capable of switching machine status rapidly. Machine learning, especially deep learning, is quickly providing new powerful tools for accelerator physicists to build fast-prediction surrogate models \cite{Gonzalez2017, Emma2018, Edelen2020} or extract essential information \cite{Ren2020, Xu2020, Tennant2020} from large amounts of data in recent years. These machine learning models can be extremely useful for building virtual accelerators which are capable of making fast predictions of the behavior of beams \cite{Nagaitsev2021}, assisting accelerator tuning by virtually bringing destructive diagnostics online \cite{Emma2018}, providing an initial guess of input parameters for a model-independent adaptive feedback control algorithm \cite{Scheinker2018, Leemann2019} or driving a model-based feedback control algorithm \cite{Emma2021}. Deep learning is a subfield of machine learning based on artificial neural networks \cite{Goodfellow2016}. One way of training a neural network model is to make use of simulated data. However, beam dynamics simulations are typically carried out under different theoretical assumptions on collective effects such as space charge forces, wakefields and coherent synchrotron radition. In addition, electron emission from a photocathode is governed by multiple physical processes and is even more difficult to simulate \cite{Moody2018}. Moreover, aging of accelerator components affects the long-term operation of a facility, but is generally not included in simulation. As a result, it is extremely challenging to achieve a good agreement between simulation and measurement for a large range of machine operation parameters even exploiting complicated physical models \cite{Chen2020}. Furthermore, it can be prohibitively expensive to collect a large amount of high-resolution simulation data \cite{Qiang2017}. 

Previous work has demonstrated prediction of the measured longitudinal phase-space at the exit of the LCLS accelerator using the L1S phase and a shallow multi-layer perceptron \cite{Emma2018}. The images were cropped to 100 x 100 pixels and the phase-space distribution must be centered in order to produce reasonable results. Nonetheless, the predicted longitudinal phase-space is blurry and has significant artifacts in the background. Moreover, the current profile was predicted by using another multi-layer perceptron instead of extracted directly from the predicted longitudinal phase-space. Indeed, a multi-layer perceptron consisting of purely fully connected layers has intrinsic limitations in image-related tasks as it intends to find the connection between each pair of nodes between each two adjacent layers. First of all, it unnecessarily complicates the training of the neural network as pixels representing the phase-space distribution apparently has little connection with majority of the background pixels. Secondly, the number of parameters scales at least proportionally to the number of pixels in the image, which makes it impractical to be applied on megapixel images due to the huge memory requirement. In \cite{Edelen2019}, convolutional and upsampling layers are used in predicting simulated longitudinal phase-spaces for the LCLS. The results do not show artifacts in the background. However, details of the study are not reported.

%
\begin{figure*}[hbt!]
	\includegraphics[width=1.0\textwidth]{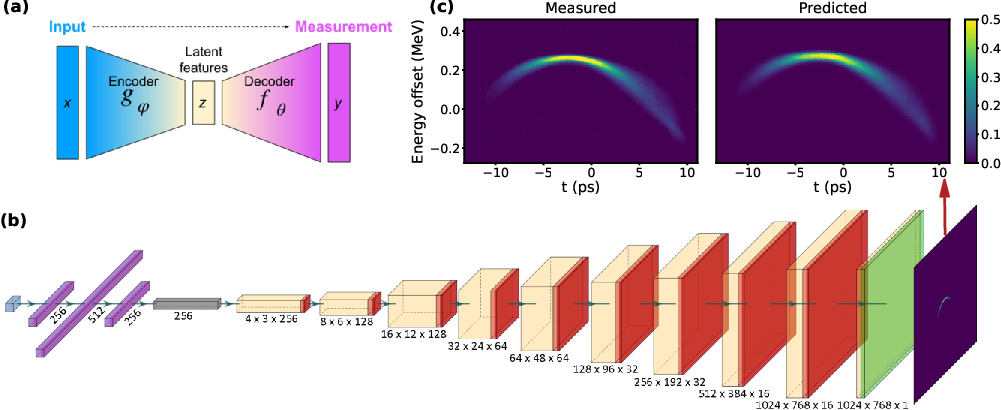}
	\caption{\label{fig:network} (a) General architecture of the encoder-decoder model. (b) Diagram of the neural network. The leftmost blue box represents the input layer. It is followed by three fully-connected layers (encoder) in purple with each layer activated by the Leaky ReLU (Rectified Linear Unit) function. The latent space is depicted in grey. The ten yellow boxes represent the transposed convolutional layers (decoder). Each transposed convolutional layer is followed by a batch normalization layer \cite{Goodfellow2016} and activated by the leaky ReLU function except the last one, which is activated by the sigmoid function depicted in green. The kernel sizes of the first and second transposed convolutional layers are 3 x 4 and 3 x 3, respectively, and the kernel sizes of the other eight transposed convolutional layers are all 5 x 5. The total number of trainable parameters is 1,898,161. (c) Example of the longitudinal phase-spaces cropped from the measured image and the corresponding prediction.}
\end{figure*}
%

In this paper, we propose an encoder-decoder neural network model to make high-fidelity predictions of megapixel images measured on a screen and demonstrate it experimentally at the longitudinal phase-space diagnostic beamline at the injector of the European XFEL. This method can provide not only high-quality virtual diagnostics while online tuning of a photoinjector but also blazingly fast machine-specific offline simulation. Besides the performance, another major advantage of this approach over the existing ones \cite{Emma2018, Emma2021} is that the output of our model is the full image from the camera. Therefore, the same neural network structure can be applied to distributions with different footprints, for example, when beam properties such as energy spread and bunch length change significantly. The concerned physical properties can then be extracted by using the well-established routines. The model learns only from the experimental data without any prior knowledge of RF photoinjectors and electron beams, which makes it potentially applicable to many other image-based diagnostics. More importantly, we demonstrate the scalability and interpretability of the model by sharing the same decoder with encoders used for different setups of the photoinjector, and propose a pragmatic way to model a photoinjector with various diagnostics and working points. It must be pointed out that our method is essentially different from the variational autoencoder \cite{Kingma2019} and the generative adversarial network \cite{Goodfellow2014}, both of which learn a joint probability distribution from the training dataset, allowing to synthesize images from random noise. In this study, however, we aim to find an explicit mapping between the input parameters and the output image.

\section{Deep Learning Model}

\subsection{Neural network}

The general architecture of the encoder-decoder model is illustrated in Fig. \ref{fig:network}(a). More generally, given an input $\mathbf{x} \in \mathbb{R}^{m}$ and the measurement $\mathbf{y} \in \mathbb{R}^{n}$, the model is asked to learn two neural networks $g_{\varphi}: \mathbb{R}^{m} \to \mathbb{R}^{c}$ and $f_{\theta}: \mathbb{R}^{c} \to \mathbb{R}^{n}$, where $\mathbb{R}^{c}$ is the latent space and $\mathbf{z} \in \mathbb{R}^{c}$ is called the latent features. Both $m$ and $n$ can be very large as modern area detectors typically have millions of pixels. The learning process is described as minimizing a loss function $\mathcal{L}(\mathbf{y}, f_{\theta}(g_{\varphi}(\mathbf{x}))$ using a gradient descent algorithm. Therefore, the model only learns from non-fixed input data $\tilde{\mathbf{x}}$ and the encoder can be simplified to $g_{\varphi}(\mathbf{x}) = g_{\varphi}(\tilde{\mathbf{x}} | \bar{\mathbf{x}}) = g_{\varphi}(\tilde{\mathbf{x}})$, where $\bar{\mathbf{x}}$ is the fixed input data and $\bar{\mathbf{x}} \oplus \tilde{\mathbf{x}} = \mathbf{x}$. Here we have assumed that the influence of the jitter of $\bar{\mathbf{x}}$ is negligible. Although it can be challenging for neural networks to learn a universal approximator for the whole input parameter space of an accelerator, this approach can be well-suited for user facilities as they are typically operated on a finite number of working points.

The detailed structure of the model is shown in Fig.~\ref{fig:network}(b). We use a multi-layer perceptron to learn latent features and then map them to the image on the screen using a concatenation of transposed convolutional layers \cite{Dumoulin2016}. The transposed convolutional layer performs the transformation in the opposite direction of a normal convolution, which projects localized feature maps to a higher-dimensional space. Despite of the deepness of the neural network, a single prediction only takes about 20 ms on a NVIDIA Tesla P100-16GB graphics card, which is orders of magnitude faster than standard beam dynamics simulation.

\subsection{Loss function}

Neural networks are trained using the mini-batch stochastic gradient decent optimization algorithm \cite{Goodfellow2016} driven by a loss function. For most of the regression problems, the choice of the loss function defaults to the mean squared error (MSE) \cite{Edelen2020, Emma2018, Ren2020, Edelen2019}. However, a MSE loss function treats pixels as uncorrelated features and was found to result in overly smoothed images as well as loss of high-frequency features in high-resolution image generation applications \cite{Ledig2017}. In our model, the loss function takes into account the correlations between adjacent pixels and is given by
\begin{equation}
L_{batch} = \frac{1}{N_b}\sum_{i=1}^{N_b}(1 - h(\mathbf{y}_{i}, \mathbf{\hat{y}_{i}})),
\end{equation}
where $N_b$ the batch size for training, $\mathbf{y}$ the measurement, $\mathbf{\hat{y}}$ the prediction and $h$ is the SSIM (structural similarity index measure) \cite{Wang2004} in multiple scales written as
\begin{widetext}
\begin{equation}
	h(\mathbf{y}, \mathbf{\hat{y}}) = \left[ \frac{1}{N_p^{(M)}} \sum_{ \substack{ \forall \mathbf{p} \in \mathbf{y}^{(M)} \\ \forall \mathbf{\hat{p}} \in\mathbf{\hat{y}}^{(M)} } } { 
		l(\mathbf{p}, \hat{\mathbf{p}})} c(\mathbf{p}, \hat{\mathbf{p}}) s(\mathbf{p}, \hat{\mathbf{p}}) \right] ^ {\alpha_M}
		\prod_{j=0}^{M-1} \left[ \frac{1}{N_p^{(j)}} \sum_{ \substack{ \forall \mathbf{p} \in \mathbf{y}^{(j)} \\ \forall \mathbf{\hat{p}} \in\mathbf{\hat{y}}^{(j)}} } c(\mathbf{p}, \hat{\mathbf{p}}) s(\mathbf{p}, \hat{\mathbf{p}}) \right] ^ {\alpha_j}.
\label{eqn:ssim}
\end{equation}
\end{widetext}
Here, $l(\mathbf{p}, \hat{\mathbf{p}})$, $c(\mathbf{p}, \hat{\mathbf{p}})$ and $s(\mathbf{p}, \hat{\mathbf{p}})$ measure the distortions in luminance, contrast and structure \cite{Wang2004}, respectively, between a uniform sliding window $\mathbf{p}$ of size 8 x 8 pixels on the measured image $\mathbf{y}^{(j)}$ and its counterpart $\hat{\mathbf{p}}$ on the predicted one $\hat{\mathbf{y}}^{(j)}$. The number of pixels in $\mathbf{y}^{(j)}$ is denoted as $N_p^{(j)}$. The superscription $j \in \{0,...,M\}$ indicates that the image is downsampled by a factor of $2^j$ using average pooling. Because $l(\mathbf{p}, \hat{\mathbf{p}})$, $c(\mathbf{p}, \hat{\mathbf{p}})$ and $s(\mathbf{p}, \hat{\mathbf{p}})$ all range between 0 and 1 for non-negative image data, having $\alpha_j < 1$ prevents the model from overfitting on fine local features which could be induced by machine jitter. Comparisons between images at different scales obviously enable the model to learn the correlations between pixels in a wider area. We empirically chose $M=2$ with $\alpha_0$ = 0.05, $\alpha_1$ = 0.30 and $\alpha_2$ = 0.65 for this study. 

\section{Experimental results}

\subsection{Experiment setup}

\begin{figure}[h]
	\includegraphics{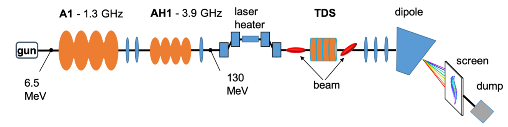}%
	\caption{\label{fig:beamline} Schematic of the European XFEL photoinjector and its diagnostic beamline. The phases of the gun, the 1.3 GHz cryomodule (A1) and the 3.9 GHz cryomodule (AH1) are used as input to predict the image on the screen. The laser heater was switched off during the experiment.}
\end{figure}

The experiment was carried out at the injector of the European XFEL \cite{Brinker2016} and the layout of the beamline is shown in Fig.~\ref{fig:beamline}. The nominal beam energy is $\sim$130 MeV which was measured at the maximum mean momentum gain (MMMG) phases of the gun and A1 as well as the zero-crossing \cite{Akre2001} phase of AH1. We refer to this working point as the reference working point and the corresponding phases as the reference phases. The bunch charge is around 250 pC. The transverse deflecting structure (TDS) and the dipole magnet were used to measure the longitudinal phase-space at a resolution of about 0.047 ps/pixel and 0.0031 MeV/pixel. We collected data for two different working points (WPs). For WP1, the phases of the gun, A1 and AH1 were uniformly sampled within $\pm$ 3 degrees, $\pm$ 6 degrees and $\pm$ 6 degrees relative to the respective reference phases. It is worth mentioning that the actual MMMG phase of A1 and the zero-crossing phase of AH1 shift as the gun phase varies due to the time of flight change. For WP2, AH1 was switched off and the gradient of A1 was reduced accordingly to keep the norminal beam energy at $\sim$130 MeV. The sample ranges of the gun and A1 phases remain the same. The inputs for WP1 and WP2 are summarized in Table~\ref{tab:input_vector}.

\begin{table}[hbt!]
\caption{\label{tab:input_vector}%
Input parameters and their ranges for the two working points. The lengths of $\tilde{\mathbf{x}}$ for WP1 and WP2 are 3 and 2, respectively.
}
\begin{ruledtabular}
\begin{tabular}{rcc}
\textrm{}&
\textrm{WP1}&
\textrm{WP2}\\
\colrule
Gun phase (deg) & -3 $\sim$ 3 & -3 $\sim$ 3 \\
A1 phase (deg) & -6 $\sim$ 6 & -6 $\sim$ 6 \\
AH1 phase (deg) & -6 $\sim$ 6 & - \\
\end{tabular}
\end{ruledtabular}
\end{table}

\begin{figure}[h]
	\includegraphics[width=0.45\textwidth]{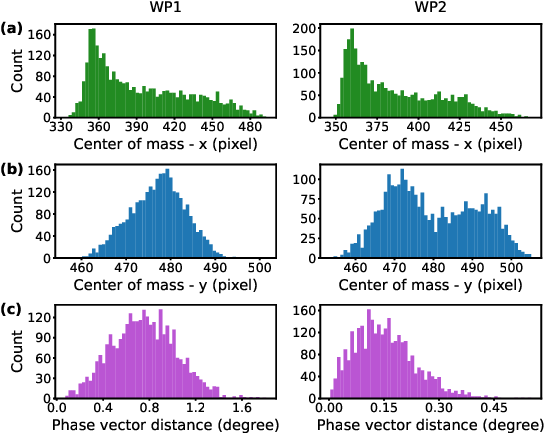}%
	\caption{\label{fig:data_quality} Statistics of the data for WP1 (the first column) and WP2 (the second column): (a-b) Histograms of the x and y coordinates of the centers of mass for the preprocessed images. (c) Histogram of the minimum Euclidean distances between the input phase vectors of each data point and the rest ones. The presence of no counts in the zero bin indicates that no data point is shared between the training and test datasets.} 
\end{figure}
%

%
\begin{figure*}[hbt!]
	\includegraphics[width=1.0\textwidth]{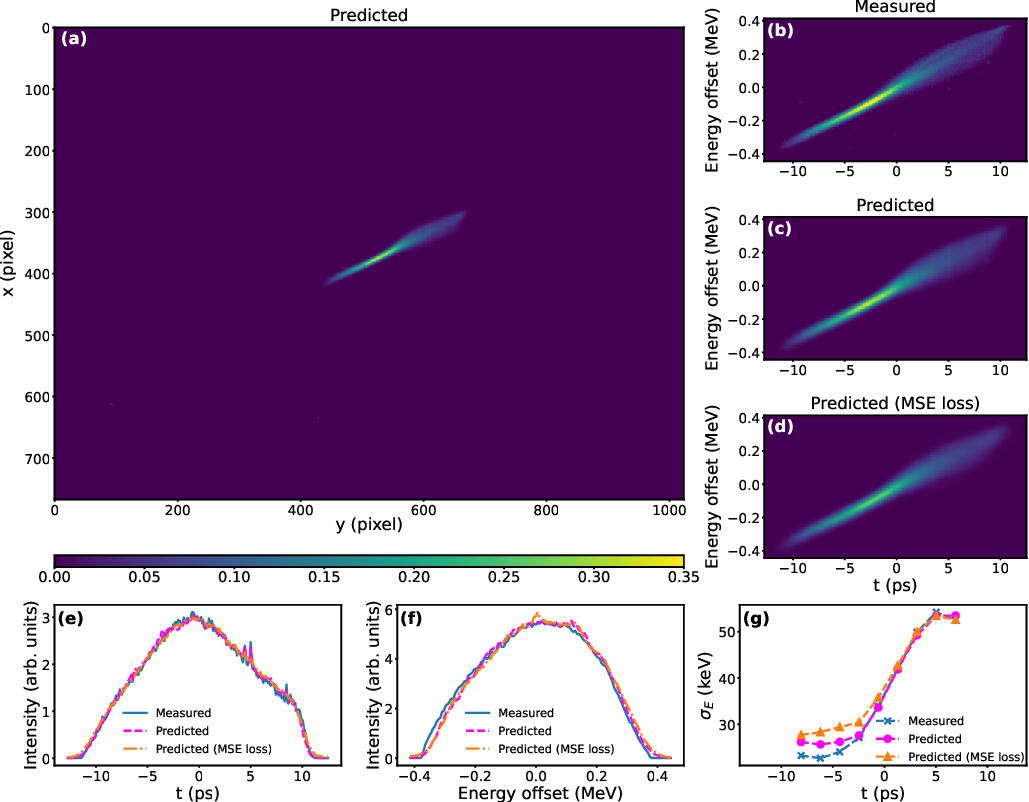}%
	\caption{\label{fig:result} (a) Example of an entire predicted image. The relative phases of gun, A1 and AH1 are -1.17 degree, -1.38 degree and 0.04 degree, respectively. (b-d) Longitudinal phase-spaces cropped from the measured image, the predicted image and the image predicted by the model using MSE as the loss function, respectively. (e-g) Comparisons of the current profiles, the energy spectra and the RMS slice energy spreads $\sigma_E$ between the longitudinal phase-spaces shown in (b-d).}
\end{figure*}
%

%
\begin{figure*}[hbt!]
	\includegraphics[width=1.0\textwidth]{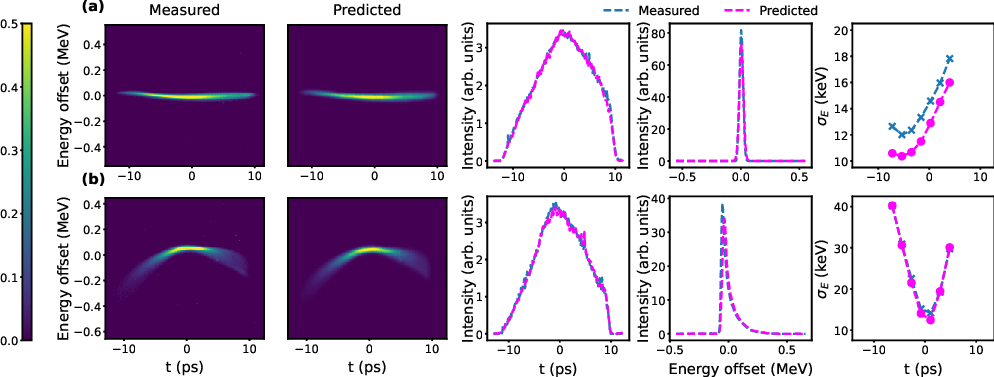}%
	\caption{\label{fig:extreme_cases} Comparisons of the measured and the predicted longitidinal phase-spaces, current profiles, energy spectra and the RMS slice energy spreads $\sigma_E$ for two shots with high peaks in the energy spectra.  (a) The relative phases of the gun, A1 and AH1 are -0.59 degree, -0.33 degree and -2.76 degree, respectively. (b) The relative phases of the gun and A1 are -2.60 degree and 0.20 degree, respectively. AH1 was switched off.}
\end{figure*}
%

%
\begin{figure*}[hbt!]
	\includegraphics[width=1.0\textwidth]{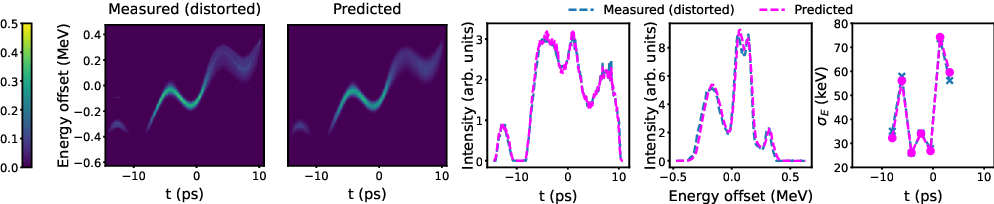}%
	\caption{\label{fig:distorted_image} Comparisons of a measured (after distortation) and the corresponding predicted longitidinal phase-spaces, current profiles, energy spectra and the RMS slice energy spreads $\sigma_E$.}
\end{figure*}
%

\subsection{Data analysis}

The original image size is 1750 x 2330 pixels. After background subtraction and normalization, all the pixel values below 0.01 were set to 0. In order to have a reasonable training time during our study with limited computational resources, all the images were slightly cropped at the same locations and then downsampled to 768 x 1024 pixels. The model was implemented and trained using the machine learning framework \textsf{TensorFlow} \cite{Tensorflow2016} version 2.3.1. For training, we adopted the weight initialization in \cite{He2015} and the Adam optimizer \cite{Kingma2014}. In total, 3,000 shots were collected for each working point. 80\% of the data were used for training and the rest were used for testing. The mini-batch size was 32 during training. The single-scale SSIM ($M=0$ and $\alpha_0$ = 1.0) is used as a metric to evaluate the performance of the trained model. It must be pointed out that this metric cannot be used an absolute quantitative measurement because low metric values can be induced by various reasons such as shift between the predicted and the measured longitudinal phase-spaces, blurry prediction and mismatched background. As mentioned previously, the proposed model does not require the phase-space distribution to be centered. Fig.~\ref{fig:data_quality}(a) and (b) show the distributions of the x and y coordinates of the centers of mass, respectively, for the preprocessed images. Evidently, the centers of mass distribute over a wide area of 160 x 46 pixels for WP1 and 122 x 52 pixels for WP2.

In machine learning, it is crucial that the information of the test dataset should not be leaked into the training dataset in order to avoid overfitting of the model. Therefore, the data points in the test dataset should not appear again in the training dataset. Fig.~\ref{fig:data_quality}(c) shows that there is no duplicated phase vector in the data for both WP1 and WP2. Therefore, randomly splitted training and test datasets will not contain the same data point.

The model for WP1 was trained with a learning rate of 1e-3 for 200 epochs and then 3e-4 for 400 epochs. In total, it took approximately 10 hours. The performance of the model over the test dataset is as high as $0.9955 \pm 0.00202$. An example predicted image is shown in Fig.~\ref{fig:result}(a). The model successfully predicts the electron distribution recorded on the screen with a clean background. The predicted longitudinal phase-space shown in Fig.~\ref{fig:result}(c) and the measured one shown in Fig.~\ref{fig:result}(b) agree very well at different longitudinal positions of the bunch, which have experienced different non-linear processes during emission from the cathode and traveling through the beamline. We also trained another model to demonstrate the influence of the loss function. The second model has the same structure as the first one but uses MSE as the loss function. The phase-space shown in Fig.~\ref{fig:result}(d) is apparently blurrier than the one shown in Fig.~\ref{fig:result}(c) although The performance of the model over the test dataset is $0.9948 \pm 0.00187$. Fig.~\ref{fig:result}(e-g) further compare the current profiles, the energy spectra and the RMS slice energy spreads of the longitudinal phase-spaces shown in Fig.~\ref{fig:result}(b-d). The predictions all agree excellently with the measurements except the slice energy spread along the first half of the bunch. Indeed, it can be distinguished from the sharpness of the image at the corresponding region. This is understandable because the input does not cover the complete state of the photoinjector. For example, the arrival time jitter of the photocathode laser \cite{Winkelmann2019} has a non-negligible impact on these regions which possess only a few pixels.

The ability of measuring high peak currents is of critical importance for a free-electron laser facility. Although all the current profiles resemble in this study, the energy spectra vary dramatically during the phase scan. Fig.~\ref{fig:extreme_cases}(a-b) show two typical results with high peaks in the energy spectra. Another model was trained on WP2 data. This model was trained with a learning rate of 1e-3 for 200 epochs and then 3e-4 for 100 epochs. The performance of the model over the test dataset is $0.9942 \pm 0.00177$. In Fig.~\ref{fig:extreme_cases}(a), the height of the peak is underestimated by about 10\% while the slice energy spread is overestimated by less than 20\%. In Fig.~\ref{fig:extreme_cases}(b), the height of the peak is underestimated by about 12\%, and the slice energy spread is only slightly overestimated at the centre of the bunch. It should be noted that the peak shown in Fig.~\ref{fig:extreme_cases}(a) is twice as high as that shown in Fig.~\ref{fig:extreme_cases}(b) due to the effect of AH1. As explained above, the precision of the model will decrease as the number of pixels which represent the distribution decreases. Nevertheless, the prediction and the measurement agree well even in these extreme cases. Consequently, it can be inferred that the model is able to predict longitudinal phase-spaces with high peak currents in the scenario where the parameter change results in a dramatical change of the current profile while the energy spectrum is stable.

\subsection{Irregular phase-space}

Despite the strong nonlinearities during the emission and transport processes of electron bunches, the shapes of the longitudinal phase-spaces in WP1 and WP2 are considered to be regular compared with those after electron bunches are compressed strongly \cite{Zhu2016} or undergo energy modulation \cite{Antipov2012}. In order to verify that the proposed model is applicable to phase-spaces with irregular shapes, the measured images of WP1 were distorted using the following equation: 

\begin{equation}
\mathbf{y}[i, j] = \mathbf{y}[i, j + 40sin(j\pi / 45)],
\label{eqn:warp}
\end{equation}
where $i$ and $j$ are the row and column indices of the image data, respectively. The model for WP1 was trained from scratch on data with the distorted images using the same procedure and hyperparameters as before. As a result, the prediction and the distorted measurement still agree excellently, as shown in Fig.\ref{fig:distorted_image}. The performance of the model over the test dataset is $0.9961 \pm 0.00199$. Although the distortion of the phase-space is not induced by nonlinear beam dynamics, it indeed reflects the generality of the neural network model.

\subsection{More on the loss function}

\begin{figure}[h!]
	\includegraphics{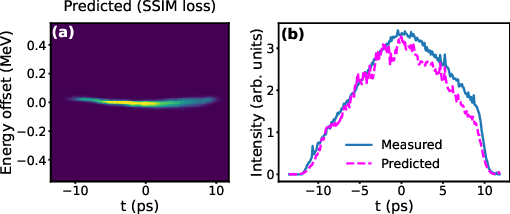}%
	\caption{\label{fig:ssim} (a) Prediction of the shot shown in Fig.~\ref{fig:extreme_cases}(a) by the model using SSIM as the loss function. (b) Comparison of the current profiles between the predicted longitudinal phase-space in (a) and the measured one shown in Fig.~\ref{fig:extreme_cases}(a).}
\end{figure}

As discussed previously, the coefficient $\alpha_j$ in Eq.~(\ref{eqn:ssim}) is critical to the performance of the model. We deliberately chose $\alpha_j < 1$ to avoid overfitting on a single scale of the image. In another word, the model is not expected to generate a precise prediction because the shot-to-shot jitter of machine parameters like the arrival time of the photocathode laser are not available as input. To illustrate the outcome of overfitting, we trained a model using the single-scale SSIM as the loss function. Namely, we ask the model to learn an exact mapping between the phase vector and the image. A typical result is shown in Fig.~\ref{fig:ssim}. The predicted longitudinal phase-space is indeed close to the measured one except that the distribution is twisted along the longitudinal axis.  Moreover, the agreement between the predicted and the measured current profiles is also not as good as the result shown in \ref{fig:extreme_cases}(a). 

The characteristics of the sliding window $\mathbf{p}$ also affects the performance of the model. The standard SSIM uses a Gaussian sliding window of size 11 x 11 pixels. It is found that the performance of the model trained with the uniform sliding window is slightly better than the one trained with the Gaussian sliding window in terms of the current profile and the energy spectra, although the latter generates a smoother image.  

%

\section{Scalability and interpretability}

\begin{figure}[h!]
	\includegraphics{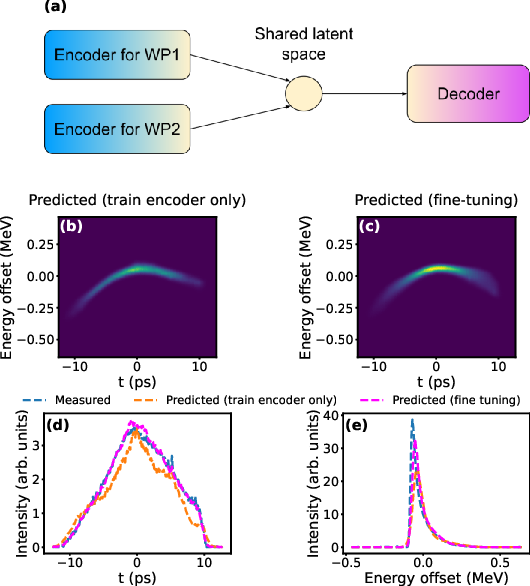}%
	\caption{\label{fig:shared_decoder} Two different encoders share a common decoder. The measured longitudinal phase-space of the example shot is shown in Fig.~\ref{fig:extreme_cases}(b). (a) General architecture of the model. (b) Prediction of a shot in WP2 after the encoder was trained together with a frozen decoder trained only on WP1 data. (c) Prediction after the decoder and both the encoders were fine-tuned. (d-e) Comparisons of the current profiles and the energy spectra between the predicted longitudinal phase-spaces in (b-c) and the measurement.}
\end{figure}

The design of the model aims at clearly separating the functions of the encoder and the decoder. Ideally, the encoder takes the input and generates the latent features which contain information about the phase-space of the electron bunch. The decoder translates the latent features into the corresponding diagnostic signal, which is the image on the screen in this study. This design leads to a scalable and interpretable model for a complex system because of parameter sharing. On the one hand, it is desirable to use the same latent features as the input for more decoders which model various diagnostics. This is also known as multi-task learning \cite{Crawshaw2020}. On the other hand, different encoders can share a common decoder, as illustrated in Fig.~\ref{fig:shared_decoder}(a), allowing for integrating multiple distinct working points into a single model. Separating the encoders for different working points is also practically necessary, because the time interval between the data collections of two working points can be significantly long so that machine parameters not used as input may have changed due to long-term phenomena such as drift.

To prove the concept of the design, we utilized the decoder trained only for the WP1 branch to train the encoder in the WP2 branch from scratch. The weights in the decoder were frozen during training. Namely, only the encoder was trained. An example of the predicted longitudinal phase-space is shown in Fig.~\ref{fig:shared_decoder}(b). Although the decoder has not experienced any data without AH1 before, the predicted longitudinal phase-space shows a qualitative agreement with the measurement. The performance of the model over the test dataset is $0.9935 \pm 0.00165$.  In order to improve the performance of the WP2 branch while preserve the performance of the WP1 branch, the model was further fine-tuned via the following steps:
\begin{enumerate}
\item  The branches for WP1 and WP2 were trained alternatedly for 100 epochs.
\item The shared decoder was frozen while the encoders for WP1 and WP2 were trained for 100 epochs.
\end{enumerate}
The above steps were repeated twice with two different learning rates: 3e-4 and 1e-4. As a result, the performance of the WP2 branch was significantly improved, as shown in Fig.~\ref{fig:shared_decoder}(c-e). The performance of the model over the test dataset increased to $0.9943 \pm 0.00177$ for WP2 and was $0.9956 \pm 0.00202$ for WP1, which are both as high as the corresponding metric achieved using the model without sharing the decoder. It should be noted that we did not optimize the above fine-tuning process and thus it could take less epochs to reach the same performance. In the long run, it is expected that the decoder will become representative enough after trained on enough data. Consequently, when a new working point is introduced, it can be required to train only a new encoder instead of the whole model with all the existing data.

Separating encoders for different working points allows to build a large model incrementally. It also helps to reduce the dimension of the input data. For example, different AH1 gradients were set for WP1 and WP2, but the AH1 gradient does not need to be included in the input. Nevertheless, a large number of varying parameters could be needed for certain applications for a larger sub-system or the whole facility. Deep learning technologies have been demonstrated to be able to solve extremely complex problems in an autonomous system provided that enough high-quality data and computational resources are available \cite{Grigorescu2020}. For scientific user facilities, the data collection speed is determined by several factors such as the repetition rate of the facility, the network delay and the relaxation time when changing a parameter as well as the performance of the data acquisition system. If the influence of long-term phenomena (e.g. drift) in a facility is negligible, the methodology of life-long learning \cite{Parisi2019} can be applied. Here, life-long learning refers to the ability of continually learn over time by accommodating new knowledge while retaining previously learned experience. For example, data can be collected from the routine operation and tuning, and then be filtered and cleaned for model training and fine-tuning. Nonetheless, considerable effort is required to build such an intelligent and robust data pipeline.

\section{Conclusion}

In summary, we have demonstrated modeling of the longitudinal phase-space diagnostic beamline at the injector of the European XFEL using encoder-decoder neural network models. After trained only with the experimental data, the model is capable of making high-fidelity predictions of megapixel images used for longitudinal phase-space measurement with RF phases as input. The prediction significantly outperforms existing methods and is orders of magnitude faster than standard beam dynamics simulation.  The longitudinal phase-space extracted from the predicted image agrees very well with the measurement not only visually, but also on important physical properties such as the current profile, the energy spectrum and the RMS slice energy spread. Due to the constraint of the computational resources, the original images were downsampled by a factor of two. This downsampling can be avoided by ultilizing a state of art graphics card or the distributed training strategy. Thus, the full-sized camera images can be used to train the model without loosing any information. In addition, a pragmatic way has been proposed to model a facility with various diagnostics and working points using deep neural networks. We have shown that the model is scalable and interpretable by sharing the same decoder with encoders used for different setups of the photoinjector. Moreover, the influences of the loss function which drives the training of the model have been discussed in depth. We conclude that the impact of the machine jitter can be mitigated by choosing proper values of the hyperparameters in the loss function at the cost of some blurring increase and accuracy loss. On the contrary, the values of the hyperparameters should be adapted to improve the accuracy of the prediction if the machine jitter is negligible.

Because both the model and the loss function do not depend on any characteristics of an RF photoinjector or the longitudinal phase-space of an electron bunch, we expect this method to be generalized to many other image-based diagnostics, not only for accelerators but also for other types of scientific facilities. The generality of the model has been demonstrated by training the model on a dataset with artificially distorted longitudinal phase-spaces. Looking forward, the model can be further extended to include more diagnostics (decoders) for the longitudinal phase-space as well as the transverse phase-space, with the ultimate goal of building a complete virtual photoinjector using experimental data.

\bibliography{pr2021zhu}

\end{document}